\documentclass[11pt]{airesearch}

\usepackage{fullpage}
\usepackage{wrapfig}

\setlist[itemize]{leftmargin=*,topsep=1pt,itemsep=1pt}

\usepackage{algorithm}
\usepackage{algpseudocode}

\usepackage{tikz}
\usetikzlibrary{arrows.meta,positioning,calc,shadows.blur}

\usepackage{fancyhdr}
\newcommand{\projectpage}[1]{%
\begin{center}
\small
\vspace{-1.25ex}
\texttt{Project Page}: \url{#1}
\end{center}
}

\ifdefined\final
\usepackage[disable]{todonotes}
\else
\usepackage[textsize=tiny]{todonotes}
\fi

\newcommand{\system}{\texttt{Lanser-CLI}}
\newcommand{\bundle}{\texttt{Analysis Bundle}}
\newcommand{\bundles}{\texttt{Analysis Bundles}}
\newcommand{\posenc}{\texttt{positionEncoding}}
\newcommand{\utf}{\texttt{UTF}}
\newcommand{\dsl}{\texttt{Selector DSL}}
\newcommand{\replay}{\texttt{Record/Replay}}
\newcommand{\rlcsf}{\textsc{RLCSF}}

\title{\huge \bfseries Reinforcement Learning from Compiler
\\and Language Server Feedback}
\author{\textbf{Yifan Zhang}\\[1.5mm]
Princeton University\\[0.5mm] \texttt{yifzhang@princeton.edu}}
\date{October 24, 2025\footnote{Revised: May 1, 2026}}

\begin{document}
\maketitle

\begin{abstract}
Coding agents fail when text-level guesses outrun program facts: they hallucinate APIs, drift to the wrong symbol, and apply edits without evidence that the workspace remains valid. Compilers, type checkers, and language servers already compute the missing supervision signal, in the form of diagnostics, symbol resolution, type information, references, and refactoring preconditions, but expose it through interfaces designed for human-driven IDEs rather than learning loops.

We introduce \textbf{Reinforcement Learning from Compiler and Language Server Feedback} (\rlcsf) together with \textbf{\system}, a CLI-first orchestration layer that exposes this signal to agents and CI. \rlcsf\ treats each tool interaction as a transition and computes a shaped process reward from deterministic changes in diagnostics, selector confidence, and edit safety. \system, in turn, converts ephemeral LSP sessions into replayable \bundles\ with pinned environment metadata and stable content hashes. Its core mechanisms are robust selectors that go beyond \texttt{file:line:col}, deterministic bundle normalization, preview-first guarded mutations, and a reward functional whose potential-based component is replayable under frozen snapshots. We formalize determinism for canonical bundles and prove that componentwise-improving transitions receive non-negative reward in the undiscounted setting. Together, these pieces yield a practical substrate for process supervision of coding agents.
\end{abstract}

\projectpage{https://github.com/yifanzhang-pro/lanser-cli}

\section{Introduction}
Large language models (LLMs) are increasingly capable coding assistants, yet their predictions about program structure, side effects, and symbol identity remain ungrounded unless checked against the actual workspace. The mismatch is especially costly for autonomous agents: a plausible edit can target the wrong definition, break a hidden type invariant, or leave diagnostics worse than before. Compilers, type checkers, and language servers already compute the facts agents need, definitions, references, types, diagnostics, and refactoring preconditions, but they were built for interactive editing rather than for optimization loops.

This raises a concrete question:

\begin{center}
\bfseries \Large {How can coding agents learn and plan from compiler \\and language server feedback?}
\end{center}

Our answer is \rlcsf. Rather than treating tool feedback as a terminal pass/fail signal, \rlcsf\ exposes dense, machine-checked feedback after each intermediate action, locating a symbol, requesting diagnostics, previewing a rename, or applying a guarded edit, and turns it into a transition-level reward suitable for planning, reinforcement learning, offline process supervision, and counterfactual evaluation.

Realizing this idea requires more than a thin wrapper around the Language Server Protocol. LSP sessions are stateful, positional, and difficult to replay; server defaults often disagree with agent-side encodings; and mutating operations can silently act on stale or ambiguous locations. \system\ addresses these issues with a CLI-first, agent-native contract that composes naturally with Unix tooling, serializes cleanly to JSONL artifacts, and is easy to containerize and gate in CI.

\begin{figure}[ht]
\centering
\begin{tikzpicture}[
  font=\footnotesize,
  node distance=6mm and 6mm,
  box/.style={draw, rounded corners=2pt, fill=blue!6, inner sep=6pt, line width=0.6pt, blur shadow, align=center},
  bigbox/.style={box, minimum width=0.7\linewidth},
  smallbox/.style={box, minimum width=0.34\linewidth},
  edge/.style={-Latex, line width=0.7pt},
  bidiredge/.style={{Latex[length=2.2mm]}-{Latex[length=2.2mm]}, line width=0.7pt}
]
\node[smallbox] (agent) {\textbf{Language Agent} \\ {\scriptsize (GPT Codex, Claude Code)}};
\node[smallbox, right=30mm of agent] (orch) {\textbf{Tool Orchestrator}\\ {\scriptsize (\system)}};

\node[bigbox, below=10mm of $(agent.south)!0.5!(orch.south)$] (ls) {\textbf{Compiler / Language Server}\\(e.g. Pyright)};
\node[bigbox, below=8mm of ls] (ws) {\textbf{Code Workspace}};

\draw[bidiredge] (agent.east) -- node[above=1.6mm] {\scriptsize Requests / Rewards} (orch.west);
\draw[edge] (orch.south) -- node[right=1mm] {\scriptsize JSON-RPC / Analyzer Adapter} (ls.north);
\draw[edge] (ls.south) -- node[right=1mm] {\scriptsize Files \& Index} (ws.north);
\end{tikzpicture}
\caption{A language agent interacts with the \system\ orchestrator, which speaks JSON-RPC to a pinned language server or compiler-backed analyzer over a concrete workspace. The orchestrator turns transient protocol sessions into stable artifacts and transition rewards.}
\end{figure}

\system\ makes compiler and language server feedback usable as a learning signal through four mechanisms:
\begin{itemize}
    \item A \dsl\ that addresses code through symbols, AST paths, content anchors, and explicit coordinate encodings, reducing reliance on brittle \texttt{file:line:col} references.
    \item \bundles\ that normalize tool responses, record environment and capability metadata, and carry stable content hashes for replay and auditing.
    \item Preview-first rename and apply flows protected by workspace jails, Git-aware staging, conflict reporting, and ambiguity checks.
    \item An \rlcsf\ reward functional that converts diagnostic deltas, selector confidence, safety readiness, and structured tool errors into a replayable transition signal.
\end{itemize}

We instantiate the system against the Language Server Protocol (LSP)~\citep{lsp-spec-3.17} using Pyright for Python~\citep{pyright}. The formal results in \Cref{sec:process-rewards} establish that bundle hashes are stable under frozen snapshots, and that the undiscounted reward is non-negative whenever every tracked component weakly improves and no tool error is raised.

\section{System Design for Feedback-Oriented Language Agents}
\label{sec:lanser-cli}

\subsection{Bridging the Agent-Feedback Gap}
Language servers were designed for interactive IDEs, not autonomous optimization loops. Operating them at agent scale raises four first-class requirements:
\begin{itemize}
    \item \emph{Determinism.} Equivalent requests should produce byte-stable artifacts after response normalization, version pinning, and content hashing.
    \item \emph{Robust addressing.} Agents need selectors that survive edits, expose ambiguity, and make encoding conventions explicit.
    \item \emph{Safe mutation.} Refactors must be previewed, confined to the workspace, checked for conflicts, and recoverable when application fails.
    \item \emph{Dense supervision.} Intermediate tool feedback should be transformed into a verifiable signal that correlates with successful repair and refactoring.
\end{itemize}

\system\ addresses these requirements by turning interactive compiler and language server sessions into verifiable artifacts. The resulting interface gives LLM agents protocol grounding: model speculation is replaced by machine-checked facts, and any pair of adjacent bundles can be scored by a deterministic reward functional.

\begin{remark}[Bootstrapping]
\label{rem:bootstrapping}
\system\ is used during its own development: we run \texttt{Lanser-CLI} to prepare and preview refactors in this repository, validate schema changes against historical traces, and replay bundles in CI to detect nondeterminism.
\end{remark}

\subsection{Architecture Overview}
At the core of \system\ is an orchestrator that mediates all agent--tool communication. It manages the language server lifecycle (start/stop, capability negotiation, cancellation, and restart with backoff), synchronizes document state, normalizes server responses, and emits \bundles. Feedback from compilers and type checkers enters either through language server diagnostics or through adapters that expose the same bundle schema.

Beyond session management, the orchestrator coalesces identical in-flight queries via a single-flight cache and serves later callers from memoized bundles. A tracing layer records JSON-RPC frames and workspace digests so that \replay\ can regenerate byte-stable outputs offline for auditing and regression testing.

\medskip\noindent\textbf{Environment capture.}
Each \bundle\ records \texttt{\{toolVersion, serverVersion, \\positionEncoding,pythonExe, pythonVersion, venvPath, configDigest, platform\}}, enabling reproducibility checks and differential debugging across machines.

\medskip\noindent\textbf{Contracts and invariants.}
All location lists are ordered by the total order $(\texttt{uri}, sL, sC, eL, eC)$ with stable tie-breakers. The \texttt{bundleId} is the SHA-256 of a JCS-canonicalized subset of fields that excludes volatile timestamps and run-local trace data. Given an identical workspace snapshot, tool version, encoding, and request, \system\ yields a byte-identical hash-domain bundle under deterministic tool semantics (see \Cref{prop:determinism}). Replayability extends to any scalar computed from canonical bundle contents and recorded parameters; in particular, transition rewards are computed from ordered pairs of adjacent bundles.

\section{Selectors and Repositioning}
\label{sec:selectors}

\subsection{The \dsl\ for Robust Addressing}
Agents need references that survive edits, but raw \texttt{file:line:col} coordinates do not. The \system\ \dsl\ captures \emph{intent} rather than absolute byte offsets by unifying several addressing strategies under a single contract. Selectors are represented programmatically as a \texttt{PositionSpec} tagged union and textually as a canonical string; both forms resolve to concrete ranges through a deterministic relocation procedure.

\medskip\noindent\textbf{PositionSpec.}
The \texttt{PositionSpec} is the internal, structured representation shared by all selector kinds:
\begin{itemize}
    \item \texttt{Cursor:} \texttt{\{kind:"cursor", uri, line, col, indexing:"utf-16|utf-8|codepoint"\}}
    \item \texttt{Range:} \texttt{\{kind:"range", uri, start:[l,c], end:[l,c]\}}
    \item \texttt{Symbolic:} \texttt{\{kind:"symbol", qualname:"pkg.mod:Class.method", role:"def|sig|body|doc", overload:0\}}
    \item \texttt{AST path:} \texttt{\{kind:"ast", path:[["module","pkg.mod"],["class","C"],["def","m"]]\}}
    \item \texttt{Content anchor:} \texttt{\{kind:"anchor", uri, snippet:"def load\_data(", \\ctx:24, hash:"sha1:..."\}}
\end{itemize}
All forms optionally carry \texttt{docVersion}, a document snapshot identifier that pins relocation to a known file version.

\medskip\noindent\textbf{Canonical string form.}
For CLI usage, logging, and human readability, the \dsl\ provides a compact, canonical syntax that maps directly to \texttt{PositionSpec} structures:
\begin{verbatim}
# Cursor/range
src/app.py@L42:C7
src/app.py@R(42,7->44,1)

# Symbolic
py://pkg.mod#Class.method:body
py://pkg.mod#function_name:sig

# AST path (subset)
ast://[module=pkg.mod]/[class=Class]/[def=method]/name[1]

# Content anchor (snippet + context N chars)
anchor://src/app.py#"def load_data("?ctx=24
\end{verbatim}

\subsection{Indexing Semantics and Encoding}
For coordinate-based selectors such as \texttt{Cursor} and \texttt{Range}, position encoding is a frequent source of off-by-one errors. \system\ negotiates \posenc\ with the server at \texttt{initialize}, preferring \texttt{utf-16} per the LSP specification while also supporting \texttt{utf-8}.

The server operates on its negotiated encoding, but CLI I/O can be declared independently via \texttt{--index-io=utf-8|utf-16|codepoint}, where \emph{codepoint} denotes Unicode scalar values. When the two indexings differ, \system\ emits both coordinate systems in verbose mode and records the server-side encoding in bundle metadata. This makes explicit, and resolves, the long-standing ambiguity between LSP's default UTF-16 indexing and the UTF-8 conventions common in downstream tools~\citep{lsp-spec-3.17}.

\subsection{Repositioning and Ambiguity Resolution}
Even with encodings resolved, selectors must remain meaningful as code evolves. The \textsc{Relocate} algorithm (\Cref{alg:relocate}) resolves a possibly stale selector against the current workspace and reports ambiguity with ranked, deterministic evidence.

\medskip\noindent\textbf{Strategy.}
\system\ first consults an exact \texttt{docVersion} map when the referenced snapshot is available. Symbolic and AST-based selectors trigger a reparse of the current workspace and a structural match. Content-anchored selectors invoke a fuzzy search using winnowed $k$-grams within the recorded context window. The resulting candidates are scored, sorted, and returned with deterministic evidence; mutating commands require explicit confirmation whenever the accepted target is ambiguous.

\medskip\noindent\textbf{Scoring.}
Exact snapshot maps and exact anchor-hash matches are treated as certified candidates and assigned score $1$; all other candidates are ranked by a deterministic score. Let $s_{\text{ast}}$ be an AST-kind match indicator, $s_{\text{module}}$ a module-equivalence score, $J_{\text{token}}$ token Jaccard, and $s_{\text{prox}}$ a proximity score. All features are normalized to $[0,1]$, with inapplicable features set to $0$ and recorded in the explanation. Non-certified candidates are ranked by the convex combination
\begin{equation}
\mathrm{score}(s,c) \;=\; 0.5\,s_{\text{ast}} \;+\; 0.2\,s_{\text{module}} \;+\; 0.2\,J_{\text{token}} \;+\; 0.1\,s_{\text{prox}}.
\end{equation}
Candidates are sorted by descending score and then by ascending $(\texttt{uri},\texttt{range})$, inducing a total order. Let $\Delta$ be the margin between the top two scores, with $\Delta=\infty$ when only a single candidate exists. A selector is accepted automatically only when the top score is at least $\tau$ and $\Delta\ge\delta$; otherwise the bundle surfaces top-$k$ alternatives with explanations, and mutating commands require explicit target confirmation.

\begin{algorithm}[t]
\caption{\system\ Repositioning (\textsc{Relocate})}
\label{alg:relocate}
\begin{algorithmic}[1]
\Require Selector $s$, workspace $W$, optional snapshot $v$, thresholds $\tau,\delta$
\Ensure Ranked candidates $\mathcal{C}$ with explanations
\State $\mathcal{C}\gets\emptyset$
\If{$v$ is present and $W$ has exact map$(s,v)$} \Return $\{(\text{map}(s,v),1.0,\text{certified})\}$ \EndIf
\If{$s.kind\in\{\text{symbol},\text{ast}\}$}
\State $\mathcal{A}\gets$ resolve\_structural$(s,W)$ \Comment{module import graph + parser}
\State $\mathcal{C}\gets\mathcal{C}\cup\mathcal{A}$
\EndIf
\If{$s.kind=\text{anchor}$}
\State $\mathcal{H}\gets$ fuzzy\_within\_ctx$(s.snippet,s.ctx; k{=}7, w{=}4)$
\State $\mathcal{C}\gets\mathcal{C}\cup\mathcal{H}$
\EndIf
\ForAll{$c\in\mathcal{C}$}
\If{$c$ is certified} \State $c.score\gets 1.0$
\Else \State $c.score\gets f(s,c)$ \Comment{Eq.~(1): deterministic weights}
\EndIf
\EndFor
\State $\mathcal{C}\gets$ sort$(\mathcal{C},-\text{score},\text{uri},\text{range})$
\If{$\mathcal{C}=\emptyset$} \State \Return \textsc{Error}(\texttt{E/NOT\_FOUND}) \EndIf
\If{$\mathcal{C}[1].score<\tau$ or $(|\mathcal{C}|>1$ and $\mathcal{C}[1].score-\mathcal{C}[2].score<\delta)$}
\State attach \texttt{disambiguation} evidence
\EndIf
\State \Return $\mathcal{C}[1..k]$
\end{algorithmic}
\end{algorithm}

\medskip\noindent\textbf{Correctness sketch.}
Under a frozen snapshot, an exact \texttt{docVersion} map returns the corresponding certified range. Symbolic and AST selectors either resolve to a unique structural target or produce a ranked candidate set with explicit disambiguation evidence; uniqueness is not assumed in the presence of overloads, shadowing, or duplicate names. For anchors, if the snippet hash and context match exactly and no conflicting exact matches exist, \textsc{Relocate} returns a certified candidate with score $1$; otherwise it ranks candidates by Eq.~(1). Deterministic sort keys ensure identical outputs across runs.

\medskip\noindent\textbf{Error taxonomy.}
Bundles carry structured error codes and, where applicable, disambiguation candidates with scores and explanations. Common errors include
\texttt{E/NOT\_FOUND}, \texttt{E/AMBIGUOUS}, \texttt{E/VERSION\_SKEW}, and \texttt{E/INDEXING\_MISMATCH}.

\section{Interfaces, Bundles, and Safety}
\system\ exposes a CLI-first interface aimed at interactive debugging, automated agent loops, and CI. The surface is organized around navigation, mutation, batch execution, and schema validation.

\paragraph{Navigation.}
Read-only commands, \texttt{lanser def}, \texttt{refs}, \texttt{hover}, \texttt{symbols}, and \texttt{diag}, all accept any \textsc{PositionSpec}. A dedicated \texttt{lanser locate} command resolves abstract selectors into concrete ranges and can preview the targeted source span, making selector intent auditable before any downstream action is taken.

\medskip\noindent\textbf{Safe mutation.}
Mutating commands default to preview. For instance, \texttt{lanser rename} is gated by \texttt{prepare-rename}: it emits a workspace edit and unified diff before any write and requires an explicit \texttt{--apply} flag to modify files. Application is protected by a workspace jail, allow/deny path filters, staged validation, conflict detection, and a clean-worktree guard that can only be overridden with \texttt{--allow-dirty}.

\medskip\noindent\textbf{Batch execution and tracing.}
\texttt{lanser batch} consumes JSONL command queues and emits JSONL bundles, supporting high-throughput planners. Any command can additionally emit an execution trace containing orchestrator metadata and JSON-RPC traffic; \texttt{lanser trace replay} uses this trace, together with a frozen workspace digest, to regenerate byte-stable outputs for auditing and regression tests.

\medskip\noindent\textbf{Schema contracts.}
\texttt{lanser schema} exports and validates JSON Schemas for selectors and bundle outputs. Agents can validate payloads before execution, and CI systems can catch incompatible schema changes before they corrupt historical traces or reward logs.

\section{Reinforcement Learning from Compiler and Language Server Feedback}
\label{sec:process-rewards}

Planner-act loops benefit from verifiable intermediate signals. In \rlcsf, the feedback source is not a human preference model but a compiler, type checker, or language server operating on the current workspace. A state contains the task, workspace snapshot, and selector context; an action is a tool query or mutation proposal; an observation is the resulting \bundle.

The reward is designed to be online-computable, replayable under frozen snapshots, and useful as \emph{shaping} rather than as a replacement for terminal task success. It follows the potential-based reward-shaping template of \citet{ng1999policy}, with the potential grounded in machine-checked program facts.

\medskip\noindent\textbf{State features.}
Let $B_t$ be the canonical bundle after step $t$, and let $\sigma_t$ be the diagnostic scope recorded in that bundle (a workspace, file, or resolved range). Let $D_t\in\mathbb{N}$ be the diagnostic count over $\sigma_t$. Diagnostic deltas are credited only when $\sigma_t=\sigma_{t-1}$; otherwise the bundle records \texttt{scope\_changed} and the diagnostic component is set to zero unless the evaluator pins a common scope. Let $S_t\in[0,1]$ denote safety readiness for a prospective mutation, with read-only steps carrying forward the previous value unless a safety check is observed. Let $A_t\in[0,1]$ denote the top selector-resolution confidence, and let $E_t\in\{0,1\}$ indicate a structured tool error such as \texttt{E/AMBIGUOUS}, \texttt{E/APPLY\_CONFLICT}, or \texttt{E/INDEXING\_MISMATCH}.

\medskip\noindent\textbf{Potential and reward.}
For non-negative weights $w_D,w_S,w_A,w_E$, define
\begin{equation}
\label{eq:potential}
\Phi(B_t) \;=\; -w_D D_t \;+\; w_S S_t \;+\; w_A A_t .
\end{equation}
The \rlcsf\ process reward is
\begin{equation}
\label{eq:proc-reward}
r^{\mathrm{csf}}_t
\;=\;
\gamma\,\Phi(B_t) - \Phi(B_{t-1}) - w_E E_t ,
\end{equation}
where $\gamma\in[0,1]$ is the RL discount used for shaping. In the common undiscounted online-planning case $\gamma=1$,
\[
r^{\mathrm{csf}}_t
=
w_D(D_{t-1}-D_t)
+ w_S(S_t-S_{t-1})
+ w_A(A_t-A_{t-1})
- w_EE_t .
\]
The reward thus credits diagnostic reduction, safety improvement, and selector-confidence improvement while penalizing structured tool failures. When an external task reward is available, an RL learner can optimize $R^{\mathrm{task}}_t+r^{\mathrm{csf}}_t$. With $w_E=0$ and bundle features included in the Markov state, the shaping term reduces to the standard potential-based form of \citet{ng1999policy}; with $w_E>0$, the tool-error penalty is an explicit task-design choice that may shift the optimal policy.

\subsection{Deterministic Analysis Bundles}
\bundles\ normalize compiler and language server payloads and pin environment metadata. Lists are deterministically ordered by $(\texttt{uri}, sL, sC, eL, eC)$ with explicit tie-breakers, and each bundle carries a stable \texttt{bundleId} computed as a hash over a canonicalized subset of its fields, excluding volatile timestamps and run-local trace data.

\paragraph{Response envelope.}
\begin{verbatim}
{
  "version": "1.2",
  "bundleId": "sha256:...",
  "status": "ok",
  "request": {"cmd": "definition", "selector": {...}},
  "resolution": {"original": "...", "resolved": {...}, "disambiguation": [...]},
  "facts": {"definitions": [...], "hover": {...}, "provenance": "lsp"},
  "edits": {"workspaceEdit": null, "diff": null},
  "processReward": {
    "version": "rl-csf-v1",
    "previousBundleId": "sha256:...",
    "r": 1.924,
    "components": {"diag_delta": 3, "safety_delta": 1,
                   "confidence_delta": 0.24, "tool_error": 0},
    "weights": {"wD":0.5,"wS":0.4,"wA":0.1,"wE":0.5,"gamma":1.0},
    "source": "compiler+lsp",
    "explanation": "Eq. (\\ref{eq:proc-reward}) over adjacent frozen bundles"
  },
  "environment": {"tool": {"name":"pyright","version":"1.1.406"},
  "positionEncoding":"utf-16","python":{"version":"3.12.0"}, ...},
  "capabilities": {"partialResult": false, "cancellable": true},
  "meta": {"exit_code": 0,
    "sorting_keys": ["uri","range[0]","range[1]","range[2]","range[3]"]
  }
}
\end{verbatim}

\begin{proposition}[Determinism under frozen snapshot]
\label{prop:determinism}
Fix a workspace snapshot $S$, a tool binary and configuration $(V,\Pi)$, a negotiated \posenc, and a request $Q$. Assume the pinned tool has deterministic semantics under these inputs, up to unordered result sets normalized by \system. Then \system\ produces identical hash-domain canonical bundles across runs; in particular, $\texttt{bundleId}(B)$ is constant. Fields explicitly excluded from the hash domain, such as timestamps and run-local trace spans, are not covered by this claim.
\end{proposition}

\begin{proof}[Proof sketch]
The orchestrator enforces deterministic sorting, canonicalizes JSON via the JSON Canonicalization Scheme (JCS)~\citep{rundgren2020rfc}, records environment invariants in the envelope, and excludes non-deterministic fields from the hash domain. Given identical inputs and deterministic tool semantics, the normalized semantic facts are a function of $(S,V,\Pi,Q)$, so the resulting hash-domain canonical JSON, and hence \texttt{bundleId}, is invariant.
\end{proof}

\begin{proposition}[Non-negativity under componentwise improvement]
\label{prop:proc-mono}
Consider Eq.~\eqref{eq:proc-reward} with $\gamma=1$ and fixed non-negative weights. If a transition weakly decreases diagnostics over the same scope ($D_t\le D_{t-1}$), weakly improves safety readiness ($S_t\ge S_{t-1}$), weakly improves selector confidence ($A_t\ge A_{t-1}$), and produces no structured tool error ($E_t=0$), then $r^{\mathrm{csf}}_t\ge0$.
\end{proposition}

\begin{proof}[Proof sketch]
For $\gamma=1$, Eq.~\eqref{eq:proc-reward} expands to $w_D(D_{t-1}-D_t)+w_S(S_t-S_{t-1})+w_A(A_t-A_{t-1})-w_EE_t$. Each difference term is non-negative by assumption and the error penalty vanishes, so the sum is non-negative. Determinism of the underlying bundles (\Cref{prop:determinism}) makes the reward replayable.
\end{proof}

\subsection{Editing and Guardrails}
\label{sec:editing-guardrails}

\system\ applies workspace edits via a staged, fail-closed workflow. Each change is written to a temporary file in the target directory while preserving detected line endings, character encoding, and (where permitted) file mode; the data is then \texttt{fsync}ed and the original is replaced via \texttt{rename(2)}. This guarantees per-file atomic replacement, but not, by itself, crash-atomicity for multi-file edits.

For multi-file edits, \system\ validates the full edit set before any replacement and records rollback metadata; whole-patch conflict detection and rollback can additionally be delegated to \texttt{git apply --3way}. Merge conflicts are surfaced as a structured \texttt{E/APPLY\_CONFLICT} carrying machine-readable hunks. Additional file-system checks are enforced as well; for example, a case-only rename such as \texttt{file.py} to \texttt{File.py} on a case-insensitive file system is rejected with \texttt{E/FS\_PERMISSIONS}.

\medskip\noindent\textbf{Threat model and safety envelope.}
Automated editing exposes several failure modes: selector resolution can target the wrong span, system failures can leave partially applied changes, writes can escape the project root, stale configuration can invalidate analysis, and encoding mismatches can corrupt positions.

\system\ counters them with a layered safety envelope. Operations are preview-by-default (\texttt{--dry-run}); a realpath-normalized workspace jail confines all file modifications to the project root, supplemented by explicit allow/deny path filters; and mutating operations require a clean Git working tree unless overridden (\texttt{--allow-dirty}). Encoding is detected automatically and, in verbose mode, dual coordinates (\utf-16 and \utf-8) are reported to prevent indexing errors. Ambiguous selectors are surfaced with confidence scores and evidence, and staged application, preflight validation, and Git-backed rollback together reduce the risk of partial failures.

Safety trade-offs are exposed as policy hooks (\texttt{--deny-apply-on-ambiguous}, \texttt{--workspace-jail}, \texttt{--allow-dirty}) so that CI systems and planning agents can configure them explicitly, making automation auditable rather than implicit.

\begin{algorithm}[t]
\caption{Guarded Rename (\textsc{PreviewThenApply})}
\label{alg:rename}
\begin{algorithmic}[1]
\Require selector $s$, new name $n$, mode $\in$ \{\texttt{dry-run}, \texttt{apply}\}
\State assert clean git worktree or \texttt{--allow-dirty}
\State \textbf{if} !\texttt{prepareRename}(s) \textbf{then} \Return \textsc{Error}
\State $E \gets$ \texttt{textDocument/rename}(s, $n$) \Comment{WorkspaceEdit preview}
\State $D \gets$ diff($E$); emit preview; \textbf{if} mode=\texttt{dry-run} \textbf{then} \Return $D$
\State stage and apply with jail + filters; \textbf{if} conflict \textbf{then} \Return \texttt{E/APPLY\_CONFLICT}
\State notify server via \texttt{didChange}; \Return success bundle with $D$
\end{algorithmic}
\end{algorithm}

These guardrails complement established program-transformation and differencing tools, e.g., GumTree~\citep{falleri2014fine} and RefactoringMiner~\citep{tsantalis2018accurate}, but emphasize determinism, auditability, and CI-grade safety envelopes.

\section{Related Work}

\noindent\textbf{Language servers, compilers, and static analysis.}
The Language Server Protocol provides a transport-agnostic interface for definitions, references, diagnostics, and edits across IDEs and tools~\citep{lsp-spec-3.17}; we instantiate \system\ with Pyright for Python~\citep{pyright}. Relative to AST differencing and refactoring systems such as GumTree~\citep{falleri2014fine} and RefactoringMiner~\citep{tsantalis2018accurate}, \system\ targets deterministic resolution, replayable artifacts, and reward construction for agent loops.

\noindent\textbf{Anchoring and robust localization.}
Content-anchored relocation in \system\ builds on local fingerprinting via winnowing~\citep{schleimer2003winnowing} and classical text-index structures such as suffix arrays~\citep{manber1993suffix}, adapted to code-aware contexts and combined with structural signals.

\noindent\textbf{Tool-using agents and process supervision.}
Language-model agents that plan and call external tools include ReAct~\citep{yao2022react}, PAL~\citep{gao2023pal}, and Toolformer~\citep{schick2023toolformer}, and step-level guidance schemes such as Self-Refine~\citep{madaan2023self} and Reflexion~\citep{shinn2023reflexion}. \rlcsf\ differs by grounding the process signal directly in compiler and language server facts and packaging those facts into deterministic bundles for replay, supervision, and credit assignment.

\noindent\textbf{Compiler feedback as reward.}
Coding agents already use tests, builds, and type checkers as terminal validators. \system\ makes this feedback finer-grained: diagnostics, selector confidence, prepare-rename checks, and apply conflicts become transition-level signals with deterministic provenance. This enables online search guidance and offline process supervision even when final success labels are sparse.

\section{Conclusion}
We presented \rlcsf\ and \system, a practical substrate for grounding coding agents in compiler and language server feedback. The underlying idea is simple: when machine-checked intermediate facts are available, agents need not learn only from terminal success or failure. Deterministic bundles make those facts replayable, robust selectors preserve intent across edits, guardrails keep mutations auditable, and the \rlcsf\ reward translates diagnostics, disambiguation confidence, and safe-apply checks into transition-level supervision. Together, these components support safer refactors, reproducible CI, and both online planning and offline credit assignment for language agents.

\vspace{5ex}
\bibliographystyle{plainnat}
\bibliography{reference}

\clearpage
\appendix

\renewcommand{\appendixpagename}{\centering \huge Appendix}
\appendixpage
\counterwithin{theorem}{section}

\startcontents[section]
\printcontents[section]{l}{1}{\setcounter{tocdepth}{2}}
\clearpage

\section{Selector Grammar and Escaping (EBNF)}
\begin{verbatim}
selector := cursor | range | symbolic | astpath | anchor
cursor := path "@" "L" INT ":" "C" INT
range := path "@" "R(" INT "," INT "->" INT "," INT ")"
symbolic := "py://" moduleref "#" qualname ( ":" role )?
moduleref:= IDENT ( "." IDENT )*
qualname := IDENT ( "." IDENT | ":" IDENT )*
role := "def" | "sig" | "body" | "doc"
path := RELPATH | "file://" URI_PATH
anchor := "anchor://" path "#" quoted_snippet ( "?" "ctx=" INT )?
quoted_snippet := '"' { char | '\"' | '\/' } '"'
\end{verbatim}
Escaping: percent-encode \texttt{\# ? \% " <space>} in anchor snippets and paths.
Windows paths canonicalize to \texttt{file:///C:/...} with an uppercase drive letter.

\paragraph{Overloads, properties, and descriptors.}
Overloaded functions can be targeted via \texttt{overload=$i$}. Properties use role \texttt{:sig} to target the getter signature; use \texttt{:def} to select the backing function object.

\section{Bundle Stability Rules}
\begin{itemize}
\item Deterministic list ordering: $(\texttt{uri}, sL, sC, eL, eC)$.
\item \texttt{bundleId} := \texttt{sha256} over a JCS-canonicalized JSON object containing \texttt{(request, resolution, facts, edits, environment, capabilities, stableMeta)}, excluding timestamps, trace spans, \texttt{processReward}, and other run-local fields.
\item Range encoding: flat \texttt{[sL,sC,eL,eC]} integer array.
\item Size limits: cap references to $10^5$ entries; mark truncation and expose a pagination cursor.
\item Canonicalization: JSON Canonicalization Scheme (JCS) with UTF-8 encoding; \texttt{meta.hashing.algo = "sha256-jcs-v1"}.
\item Dual coordinates: when CLI I/O differs from server encoding, include both coordinate systems in verbose traces; bundles retain server coordinates.
\item Reward reproducibility: \texttt{processReward} is computed from the current and previous canonical hash-domain bundle contents plus recorded weights, and is intentionally outside the current bundle's hash domain.
\end{itemize}

\section{Exit Codes}
\label{sec:appendix-exit-codes}
\begin{table}[ht]
\centering
\small
\begin{tabular}{@{}rlll@{}}
\toprule
Code & Symbol & Meaning & Retryable \\
\midrule
0& \texttt{OK} & Success & ,  \\
2& \texttt{E/BAD\_SELECTOR\_SYNTAX}& Selector parse error & No \\
3& \texttt{E/NOT\_FOUND} & No resolvable target & Sometimes \\
4& \texttt{E/AMBIGUOUS}& Multiple candidates & Yes \\
10& \texttt{E/VERSION\_SKEW}& Snapshot mismatch & Yes \\
64& \texttt{E/LS\_TIMEOUT}& Server timeout & Yes \\
65& \texttt{E/LS\_CRASH}& Server crashed & Yes \\
70& \texttt{E/APPLY\_CONFLICT}& Patch could not be applied & Manual \\
71& \texttt{E/FS\_PERMISSIONS}& Write denied & No \\
72& \texttt{E/UNSUPPORTED\_CAP} & Server lacks capability & No \\
73& \texttt{E/REQUEST\_CANCELLED} & Request was cancelled & Yes \\
74& \texttt{E/CONTENT\_MODIFIED}& Content changed mid-request & Yes \\
75& \texttt{E/INDEXING\_UNSUPPORTED}& IO indexing unsupported & No \\
76& \texttt{E/REPLAY\_MISMATCH} & Trace/workspace digest mismatch & No \\
\bottomrule
\end{tabular}
\end{table}

\section{Worked Example}

\paragraph{Definition query.}
\begin{verbatim}
lanser def py://pkg.mod#Class.method:sig --json
\end{verbatim}
Returns a \bundle\ with the resolved range, hover signature, and environment metadata such as \texttt{toolVersion=pyright@1.1.406} and \texttt{positionEncoding=utf-16}.

\paragraph{Diagnostics and reward state.}
\begin{verbatim}
lanser diag py://pkg.mod#Class.method:body --json
lanser locate py://pkg.mod#Class.method:body --json
\end{verbatim}
The resulting bundles record diagnostic counts, selector confidence, structured errors, and provenance needed to compute the \rlcsf\ transition reward.

\paragraph{Rename.}
\begin{verbatim}
lanser prepare-rename py://pkg.mod#load_data:def --json
lanser rename py://pkg.mod#load_data:def read_data --dry-run
lanser rename py://pkg.mod#load_data:def read_data --apply
\end{verbatim}
The preview includes a unified diff, the apply path enforces workspace jail and dirty-repo policies.

\section{\rlcsf\ Reward Signals: Worked Examples}

We instantiate Eq.~\eqref{eq:proc-reward} with $\gamma=1$ and
$(w_D,w_S,w_A,w_E)=(0.5,0.4,0.1,0.5)$.

\medskip\noindent\textbf{Example: Diagnostic reduction, safe apply, confident resolution.}
An agent proposes to rename \texttt{load\_data} to \texttt{read\_data}. Pyright reduces relevant diagnostics from $D_{t-1}{=}5$ to $D_t{=}2$ after a dry-run, safety readiness improves from $S_{t-1}{=}0$ to $S_t{=}1$, selector confidence improves from $A_{t-1}{=}0.70$ to $A_t{=}0.94$, and no structured tool error occurs ($E_t{=}0$). Then
\[
r^{\mathrm{csf}}_t
= 0.5\cdot(5-2) + 0.4\cdot(1-0) + 0.1\cdot(0.94-0.70) - 0.5\cdot0
= 1.924.
\]
The bundle records \texttt{\{"diag\_delta": 3, "safety\_delta": 1, "confidence\_delta": 0.24, "tool\_error": 0\}}.

\medskip\noindent\textbf{Example: Ambiguous selector and apply conflict.}
The agent attempts a refactor with unresolved imports. Diagnostics stagnate ($D_{t-1}{=}7$, $D_t{=}7$), safety readiness does not improve ($S_{t-1}{=}0$, $S_t{=}0$), selector confidence remains low ($A_{t-1}{=}0.62$, $A_t{=}0.62$), and the preview reports \texttt{E/APPLY\_CONFLICT} ($E_t{=}1$). Then
\[
r^{\mathrm{csf}}_t
=
0.5\cdot0+0.4\cdot0+0.1\cdot0-0.5\cdot1
=
-0.5,
\]
discouraging application until ambiguity and conflicts are resolved.

\medskip\noindent\textbf{Replayability.}
Because \texttt{processReward} is computed from adjacent deterministic bundle contents and fixed weights, the same $r^{\mathrm{csf}}_t$ is recovered by \texttt{lanser trace replay}. This supports offline evaluation and counterfactual policy analysis without re-running the language server.

\medskip\noindent\textbf{Design note.}
The reward is shaping, not a replacement for task success metrics. It is intended for online guidance and offline process supervision, and is non-negative under \Cref{prop:proc-mono} when the stated invariants hold.


\end{document}